\documentclass[letterpaper]{article} 
\usepackage{aaai25}  
\usepackage{times}  
\usepackage{helvet}  
\usepackage{courier}  
\usepackage[hyphens]{url}  
\usepackage{graphicx} 
\usepackage{natbib}  
\usepackage{caption} 
\usepackage{algorithm}
\usepackage{algorithmic}
\usepackage{multirow}
\usepackage{amsmath}
\usepackage{booktabs} 
\newcommand{\argmin}{\mathop{\mathrm{argmin}}}
\usepackage{newfloat}
\usepackage{listings}
\DeclareCaptionStyle{ruled}{labelfont=normalfont,labelsep=colon,strut=off} 
\floatstyle{ruled}
\newfloat{listing}{tb}{lst}{}
\floatname{listing}{Listing}
\title{Query Quantized Neural SLAM}
\author {
    Sijia Jiang\textsuperscript{\rm },
    Jing Hua\textsuperscript{\rm },
    Zhizhong Han\textsuperscript{\rm }
}
\affiliations {
    \textsuperscript{\rm }Department of Computer Science, Wayne State University, Detroit, MI, USA\\
    sijiajiang@wayne.edu, jinghua@wayne.edu, h312h@wayne.edu

}

\usepackage{bibentry}

\begin{document}

\maketitle

\begin{abstract}
Neural implicit representations have shown remarkable abilities in jointly modeling geometry, color, and camera poses in simultaneous localization and mapping (SLAM). Current methods use coordinates, positional encodings, or other geometry features as input to query neural implicit functions for signed distances and color which produce rendering errors to drive the optimization in overfitting image observations. However, due to the run time efficiency requirement in SLAM systems, we are merely allowed to conduct optimization on each frame in few iterations, which is far from enough for neural networks to overfit these queries. The underfitting usually results in severe drifts in camera tracking and artifacts in reconstruction. To resolve this issue, we propose query quantized neural SLAM which uses quantized queries to reduce variations of input for much easier and faster overfitting a frame. To this end, we quantize a query into a discrete representation with a set of codes, and only allow neural networks to observe a finite number of variations. This allows neural networks to become increasingly familiar with these codes after overfitting more and more previous frames. Moreover, we also introduce novel initialization, losses, and argumentation to stabilize the optimization with significant uncertainty in the early optimization stage, constrain the optimization space, and estimate camera poses more accurately. We justify the effectiveness of each design and report visual and numerical comparisons on widely used benchmarks to show our superiority over the latest methods in both reconstruction and camera tracking. Our code is available at \url{https://github.com/MachinePerceptionLab/QQ-SLAM}.
\end{abstract}

%

\section{Introduction}
Neural implicit representations have made huge progress in simultaneous localization and mapping (SLAM)~\cite{Zhu2021NICESLAM,nicerslam,wang2023coslam,sucar2021imap,stier2023finerecon}. These methods represent geometry and color as continuous functions to reconstruct smooth surfaces and render plausible novel views, which shows advantages over point clouds in classic SLAM systems~\cite{koestler2022tandem}. Current methods learn neural implicits in a scene by rendering them into RGBD images through volume rendering and minimizing the rendering errors to ground truth observations. To render color~\cite{neuslingjie}, depth~\cite{Yu2022MonoSDF}, or normal~\cite{wang2022neuris} at a pixel, we query neural implicit representations for signed distances or occupancy labels and color at points sampled along a ray, which are integrated based on volume rendering equations.

We usually use coordinates, positional encodings, or other features as the input of neural implicit representations~\cite{Peng2020ECCV,mueller2022instant,rosu2023permutosdf,li2023neuralangelo}, which we call a query. Queries are continuous vectors which allow neural networks to generalize well on unseen queries that are similar to the ones seen before. Continuity is good for generalization but also brings huge variations for neural networks to overfit. Neural networks need to see these queries or similar ones lots of times so that they can infer and remember attributes like geometry and color at these queries, which takes significant time. However, this runtime efficiency does not meet the requirement of SLAM systems. What is more critical is that we are only allowed to conduct optimization on current frame in merely few iterations, and no beyond frames are observable.

Underfitting on these queries results in huge drifts in camera tracking and artifacts on reconstructions. Therefore, how to query neural implicit representations to make overfitting more efficiently in SLAM is still a challenge.

To overcome this challenge, we introduce query quantized neural SLAM to jointly model geometry, color, and camera poses from RGBD images. We learn a neural singed distance function (SDF) to represent geometry in a scene through rendering the SDF with a color function to overfit image observations. We propose to quantize a query into a discrete representation with a set of codes, and use the discrete query as the input of neural SDF, which significantly reduces the variations of queries and improves the performance of reconstruction and camera tracking. Our approach is to make neural networks become increasingly familiar to these quantized queries after overfitting more and more previous frames, which leads to faster and easier convergence at each frame. We provide a thorough solution to discretize queries like coordinates, positional encodings, or other geometry features for overfitting each frame more effectively. Moreover, to support our quantized queries, we also introduce novel initialization, losses, and augmentation to stabilize the optimization with huge uncertainty in the very beginning, constrain the optimization space, and estimate camera poses more accurately. We evaluate our methods on widely used benchmarks containing synthetic data and real scans. Our numerical and visual comparisons justify the effectiveness of our modules, and show superiorities over the latest methods in terms of accuracy in scene reconstruction and camera tracking. Our contributions are summarized below.

\begin{enumerate}
\item We present query quantized neural SLAM for joint scene reconstruction and camera tracking from RGBD images. We justify the idea of improving SLAM performance by reducing query variations through quantization.
\item We present novel initialization, losses, and augmentation to stabilize the optimization. We show that the stabilization is the key to make quantized queries work in SLAM.
\item We report state-of-the-art performance in scene reconstruction and camera tracking in SLAM.
\end{enumerate}

\section{Related Work}
Neural implicit representations achieve impressive results in various applications~\cite{guo2022manhattan,rosu2023permutosdf,li2023neuralangelo,mueller2022instant,Baoruicvpr2023,udiff,localn2nm2024,zhou2023levelset,multigrid}. With supervision from 3D annotations~\cite{Liu2021MLS,tang2021sign,localn2nm2024}, point clouds~\cite{atzmon2020sald,chaompi2022,ChaoSparse,Baoruicvpr2023,chao2023gridpull}, or multi-view images~\cite{GEOnEUS2022,guo2022manhattan,zhang2024learning,zhang2024gspull}, neural SDFs or occupancy functions can be estimated using additional constraints or volume rendering.

\noindent\textbf{Multi-view Reconstruction. }
Classic multi-view stereo (MVS)~\cite{schoenberger2016sfm,schoenberger2016mvs} uses photo consistency to estimate depth maps but struggles with large viewpoint variations and complex lighting. Space carving~\cite{273735visualhull} reconstructs 3D structures as voxel grids without relying on color.  

Recent methods leverage neural networks to predict depth maps using depth supervision~\cite{yao2018mvsnet,koestler2022tandem} or multi-view photo consistency~\cite{dblp:conf/cvpr/ZhouBSL17}.  

Neural implicit representations have gained popularity for learning 3D geometry from multi-view images. Early works compared rendered outputs to masked input segments using differentiable surface renderers~\cite{DVRcvpr,sun2021neucon}. DVR~\cite{DVRcvpr} and IDR~\cite{yariv2020multiview} model radiance near surfaces for rendering.  

NeRF~\cite{mildenhall2020nerf} and its variants~\cite{park2021nerfies,mueller2022instant,sun2021neucon} combine geometry and color via volume rendering, excelling in novel view synthesis without masks. UNISURF~\cite{Oechsle2021ICCV} and NeuS~\cite{neuslingjie} improve on this by rendering occupancy functions and SDFs. Further advancements integrate depth~\cite{Yu2022MonoSDF,Azinovic_2022_CVPR,Zhu2021NICESLAM}, normals~\cite{wang2022neuris,guo2022manhattan}, and multi-view consistency~\cite{GEOnEUS2022} to enhance accuracy. Depth images play a key role by guiding ray sampling~\cite{Yu2022MonoSDF} or providing rendering supervision~\cite{Yu2022MonoSDF,lee2023fastsurf}, enabling more precise surface estimation.

\noindent\textbf{Neural SLAM. }Early work employed neural networks to learn policies for exploring 3D environments. More recent methods~\cite{zhang2023goslam,tofslam,teigen2023rgb,uncleslam2023} learn neural implicit representations from RGBD images. iMAP~\cite{sucar2021imap} uses an MLP as the only scene representation in a realtime SLAM system. NICE-SLAM~\cite{Zhu2021NICESLAM} presents a hierarchical scene representation to reconstruct large scenes with more details. Its following work NICER-SLAM~\cite{nicerslam} uses monocular geometric cues instead of depth images as supervision. Co-SLAM~\cite{wang2023coslam} jointly uses coordinate and sparse parametric encodings to learn neural implicit functions. Segmentation priors~\cite{kong2023vmap,haghighi2023neural} also show their ability to improve the performance of SLAM. Also, vMAP~\cite{kong2023vmap} represents each object in the scene as a neural implicit in a SLAM system. Depth fusion is also integrated with neural SDFs as a prior for more accurate geometry modeling in SLAM~\cite{Hu2023LNI-ADFP}.

\noindent\textbf{Neural Representations with Vector Quantization. }Vector quantization, first introduced in VQ-VAE~\cite{oord2017neural} for image generation, has been applied in binary neural networks~\cite{gordon2023quantizing}, data augmentation~\cite{wu2022randomized}, compression~\cite{dupont2022coin++}, novel view synthesis~\cite{yang2023vq}, point cloud completion~\cite{fei2022vq}, image synthesis~\cite{gu2022vector}, and 3D reconstruction/generation using Transformers or diffusion models~\cite{corona2023unaligned,li2023generalized}. Unlike these approaches, we quantize input queries to approximate continuous representations for SLAM systems, addressing runtime efficiency and visibility constraints during optimization. Unlike Gaussian splatting-based SLAM methods~\cite{keetha2024splatam,MatsukiCVPR2024,hhuang2024photoslam,Huang_2024,yu2024gaussianopacityfieldsefficient}, our approach focuses on recovering high-fidelity SDFs.

\begin{figure*}
  \centering
   \includegraphics[width=0.95\linewidth]{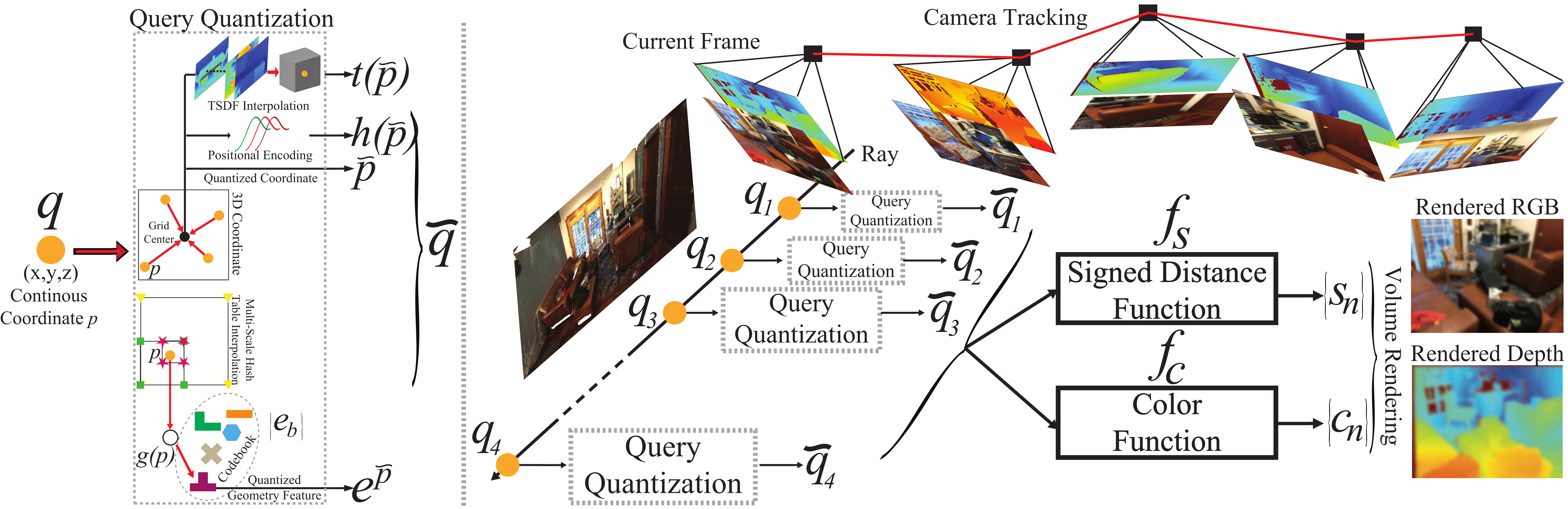}
   \caption{Overview of our method. We first quantize continuous queries $q$ into $\tilde{q}$ (left) which are then leveraged in volume rendering for camera tracking and scene mapping.}
   \label{fig:overview}
   \vspace{-0.13in}
\end{figure*}

\section{Method}
\noindent\textbf{Overview. }Following previous methods~\cite{wang2023coslam,Zhu2021NICESLAM,Hu2023LNI-ADFP}, our neural SLAM jointly estimates geometry, color and camera poses from $J$ frames of RGBD images $I$ and $D$. Our SLAM estimates camera poses $O_j$ for each frame $j$, and infers an SDF $f_s$ and a color function $f_c$ which predict a signed distance $s=f_s(\tilde{q})$ and a color $c=f_c(\tilde{q})$ for an quantized query $\tilde{q}$. $\tilde{q}$ is quantized from its continuous representation $q$, which is not limited to a coordinate $p$ but also includes its positional encoding $h(p)$, geometry feature $g(p)$, and interpolation from fused depth prior $t(p)$.

Fig.~\ref{fig:overview} illustrates our framework. Starting from a continuous query $q$, we first quantize it into a quantized query $\tilde{q}$, which is the input to our neural implicit representations including SDF $f_s$ and color function $f_c$, predicting a signed distance $s$ and a color $c$. We accumulate signed distances and colors at queries sampled along a ray into a rendered color and a depth through volume rendering. We tune $f_s$, $f_c$, and $\{O_j\}$ by minimizing rendering errors. After the optimization, we extract the zero-level set of $f_s$ as the surface of the scene using the marching cubes algorithm~\cite{Lorensen87marchingcubes}.

\noindent\textbf{Coordinate Quantization. }For the coordinate $p$ of a query $q$, we directly discretize $p$ as its nearest vertex on an extremely high resolution 3D grid, such as $12800^3$, which becomes a quantized coordinate $\tilde{p}$. We use coordinate quantization~\cite{sijia2023quantized} to reduce the coordinate variations, preserve contentiousness, and stabilize the optimization with high frequency postional encodings. Moreover, we use one-blob encoding~\cite{muller2019neural} along with the quantized coordinates $\tilde{p}$ as the positional encoding $h(\tilde{p})$. We denote $h(\tilde{p})$ as $h^{\tilde{p}}$ for simplicity in the following.

\noindent\textbf{Geometry Feature Quantization. }We follow InstantNGP~\cite{mueller2022instant} to build up a multi-resolution hash-based feature grid $\theta_g$ as geometry features in the scene. We put learnable features at vertices on the multi-resolution grid, and use the trilinear interpolation to get a geometry feature $g(p)$ at the location $p$ of query $q$. We normalize the length of $g(p)$ to be $1$ to balance the importance of different features that are used to update the same discrete code.

Following VQ-VAE~\cite{oord2017neural}, we maintain a set of $B$ learnable codes $\{e_b\}_{b=1}^{b=B}$, and quantize each geometry feature $g(p)$ into one of the codes by the nearest neighbor search using a $L_2$ norm as a metric,

\begin{equation}
\label{eq:codequantization}
e^{\tilde{p}}=\argmin\nolimits_{\{e_b\}} ||e_b-g(p)||_2^2,
\end{equation}

\noindent where we denote the nearest code to $g(p)$ in the codebook as $e^{\tilde{p}}$. After each iteration, we normalize the length of each code to be $1$ to make codes comparable with each other.

\noindent\textbf{Codebook Initialization. }Our preliminary results show that the initialization of $B$ codes really matters. Different from using relatively clean point clouds as supervision~\cite{yang2023neural}, random initialization using uniform or Gaussian distributions for each code brings more uncertainties when there are already lots uncertainties in SDF $f_s$, color function $f_c$, and the estimated camera poses in the very beginning. These uncertainties cause unstable optimization which results in large drifts in camera tracking that is hard to be corrected in the following optimization iterations. We found that using Bernoulli distribution to initialize entries to either $0$ or $1$ in each code can significantly constrain changes on these codes, and stabilize the optimization,

\begin{equation}
\label{eq:codeinit}
e_b \sim Bernoulli (0.5).
\end{equation}

\noindent\textbf{Quantization of Additional Geometry Priors. }It has shown that using additional geometry priors as a part of input can improve the reconstruction accuracy in SLAM~\cite{Hu2023LNI-ADFP}. It uses a signed distance $t(p)$ at location $p$ as a part of input. $t(p)$ is a scalar interpolated from a TSDF grid $\theta_t$ which is incrementally fused from input depth images. We simply quantize $t(p)$ by using quantized coordinates $\tilde{p}$ for the interpolation from $\theta_t$. The quantized signed distance interpolation is denoted as $t^{\tilde{p}}$.

\noindent\textbf{Quantized Queries. }To sum up, for a continuous query $q$ formed by coordinate $p$, positional encoding $h(p)$, geometry feature $g(p)$, and TSDF interpolation $t(p)$, we quantize $q$ into a discrete representation,

\begin{equation}
\label{eq:quantization}
\tilde{q}=[\tilde{p},h^{\tilde{p}},e^{\tilde{p}},t^{\tilde{p}}].
\end{equation}

\noindent\textbf{Volume Rendering. }We follow NeRF to do volume rendering at current frame $j$, we render a RGB image $\bar{I}_j$ and a depth image $\bar{D}_j$. This produces rendering errors in terms of RGB color and depth to the input $I_j$ and $D_j$, which drives the optimization to minimize.

With the estimated camera poses $O_j$, we shoot a ray $R_k$ at a randomly sampled pixel on view $I_j$. $R_k$ starts from the camera origin $o$ and points a direction of $r$. We sample $N$ points along the ray $R_k$ using stratified sampling and uniformly sample near the depth, where each point is sampled at $p_n=o+d_nr$ and $d_n$ corresponds to the depth value of $p_n$ on the ray, where each location $p_n$ indicates a query $q_n$. We quantize each query $q_n$ into $\tilde{q}_n$ using Eq.~\ref{eq:quantization}. Then, the SDF $f_s$ and the color function $f_c$ predict a signed distance $s_n=f_s(\tilde{q})$ and a color $c_n=f_c(\tilde{q})$. Following NeuralRGBD~\cite{Azinovic_2022_CVPR}, we use a simple bell-shaped function formed by the product of two Sigmoid functions $\delta$ to transform signed distances $s_n$ into volume density $w_n$,

\begin{equation}
\label{eq:rendering}
w_n=\delta(s_n/t)\delta(-s_n/t),
\end{equation}

\noindent where $t$ is the truncation distance. With $w_n$, we render RGB $\bar{I}_j$ and depth $\bar{D}_j$ images through alpha blending,

\begin{equation}
\begin{split}
\label{eq:volumerendering}
\bar{I}_j(k)=\frac{1}{\sum_{n'=1}^{N} w_{n'}}\sum\nolimits_{n'=1}^{N} w_{n'} {c}_{n'},\\
\ \bar{D}_j(k)=\frac{1}{\sum_{n'=1}^{N} w_{n'}}\sum\nolimits_{n'=1}^{N} w_{n'} d_{n'}.
\end{split}
\end{equation}

\noindent\textbf{Loss Function. }With estimated camera poses, we evaluate the rendering errors at $K$ rays on the rendered $\bar{I}_j$ and $\bar{D}_j$,

\begin{equation}
\begin{split}
\label{eq:renderingerros}
L_I&=\frac{1}{JK}\sum\nolimits_{j,k=1}^{J,K}||I_j(k)-\bar{I}_j(k)||_2^2, \\ 
 L_D&=\frac{1}{JK}\sum\nolimits_{j,k=1}^{J,K}||D_j(k)-\bar{D}_j(k)||_2^2.
\end{split}
\end{equation}

With the input depth $D_j$, we can also impose two constraints on the predicted signed distances in the free space between the camera and the surface and area near the surface. We use a threshold $tr$ of signed distances to set up a bandwidth around a surface. For queries outside the bandwidth, we truncate their signed distances into either $1$ or $-1$. Thus, an empty space loss $L_{s'}$ is used to supervise the predicted signed distances $L_{s'}=\sum_{n,k,j}||s_n-t_r||_2^2$. Moreover, we approximate the signed distances at queries $q_n$ within the bandwidth as $d_n-d_n'$, where $d_n$ is the depth observation at the pixel on $D_j$ and $d_n'$ is the depth at query $q_n$. Thus, $L_{s}=\sum_{n,k,j}||s_n-(d_n-d_n')||_2^2$ can be used to supervise the predicted signed distances $s_n$.

To learn the $B$ codes $\{e_b\}$, we impose two constraints. One is that we push the code $e^{\tilde{p}}$ that the geometry feature $g(p)$ matches in Eq.~\ref{eq:codequantization} to be similar to $g(p)$. We use a MSE,

\begin{equation}
\begin{split}
\label{eq:stopgradient}
L_{g}=||sg[e^{\tilde{p}}]-g(p)||_2 + \lambda||e^{\tilde{p}}-sg[g(p)]||_2,
\end{split}
\end{equation}

\noindent where $sg$ stands for the stop gradient~\cite{oord2017neural} operator. The key idea behind stop gradient is to decouple the training of the SDF $f_s$, color function $f_c$ from the training of $B$ codes. We use $\lambda = 0.1$
in all our experiments.The other is that we diversify the $B$ codes $\{e_b\}$ to prevent them from going to the same point in the feature space using a diverse loss $L_{e}=\sum_b^B\sum_{b'}^B||e_b-e_{b'}||_2$. 

Our loss function $L$ includes all loss terms above. We jointly minimize all loss terms with balance weights $\alpha$, $\beta$, $\gamma$, $\zeta$ and $\eta$ below,

\begin{equation}
\label{eq:loss} 
\min_{f_s,f_c,\{e_b\},\theta_g}  L_I + \alpha L_D + \beta L_{g} - \gamma L_{e} + \zeta \ L_{s} + \eta L_{s'}.
\end{equation}

\noindent\textbf{Details in SLAM. }With RGBD input, we jointly estimate camera poses for each frame and infer the SDF $f_s$ to model geometry. For camera tracking, we first initialize the pose of current frame using a constant speed assumption, which provides us a coarse pose estimation according to poses of previous frames. We use the coarse pose estimation to shoot rays and render RGB and depth images. We minimize the same loss function in Eq.~\ref{eq:loss} by only refining the camera poses and keeping other parameters fixed. We refine camera poses and other parameters at the same time in a bundle adjustment procedure  every $5$ frames, where we also add the pose of current frame as one additional parameter in Eq.~\ref{eq:loss}. 

For reconstruction, we render rays from the current view and key frames in each batch. Instead of key frame images, we follow Co-SLAM~\cite{wang2023coslam} to store rays randomly sampled $5\%$ of all pixels from each key frame in a key frame ray list. This allows us to insert new key frames more
frequently and maintain a larger key frame coverage. We select a key frame very 5 frames.

With the estimated camera poses, we incrementally fuse input depth images $D_j$ into a TSDF grid $\theta_t$ in a resolution of $256$. We do trilinear interpolation on $\theta_t$ to obtain the prior interpolation t(p) at a query q.


\noindent\textbf{Augmentation of Geometry Priors. }Although depth fusion priors~\cite{Hu2023LNI-ADFP} show that the TSDF $\theta_t$ can improve the reconstruction accuracy in SLAM, we found that the interpolation $t^{\tilde{p}}$ of geometry priors significantly degenerate the performance in our preliminary results. Our analysis shows that the neural networks learn a shortcut from the input to the output, which directly maps the geometry prior $t^{\tilde{p}}$ as the predicted signed distance at most queries, ignoring any geometry constraints like camera poses. The reason why it works well with depth fusion priors~\cite{Hu2023LNI-ADFP} is that it predicts occupancy probabilities but not signed distances, which differentiates the input from the output.

To resolve this problem, we introduce a simple augmentation to manipulate the geometry prior interpolation $t^{\tilde{p}}$ through a linear transformation. We use $t^{\tilde{p}} \leftarrow tanh(t^{\tilde{p}})$ to make signed distances still comparable to each other in the range of $[-1,1]$ but also shift away from the original TSDF interpolations.

\begin{figure}[tb]
\centering
   \includegraphics[width=0.9\linewidth]{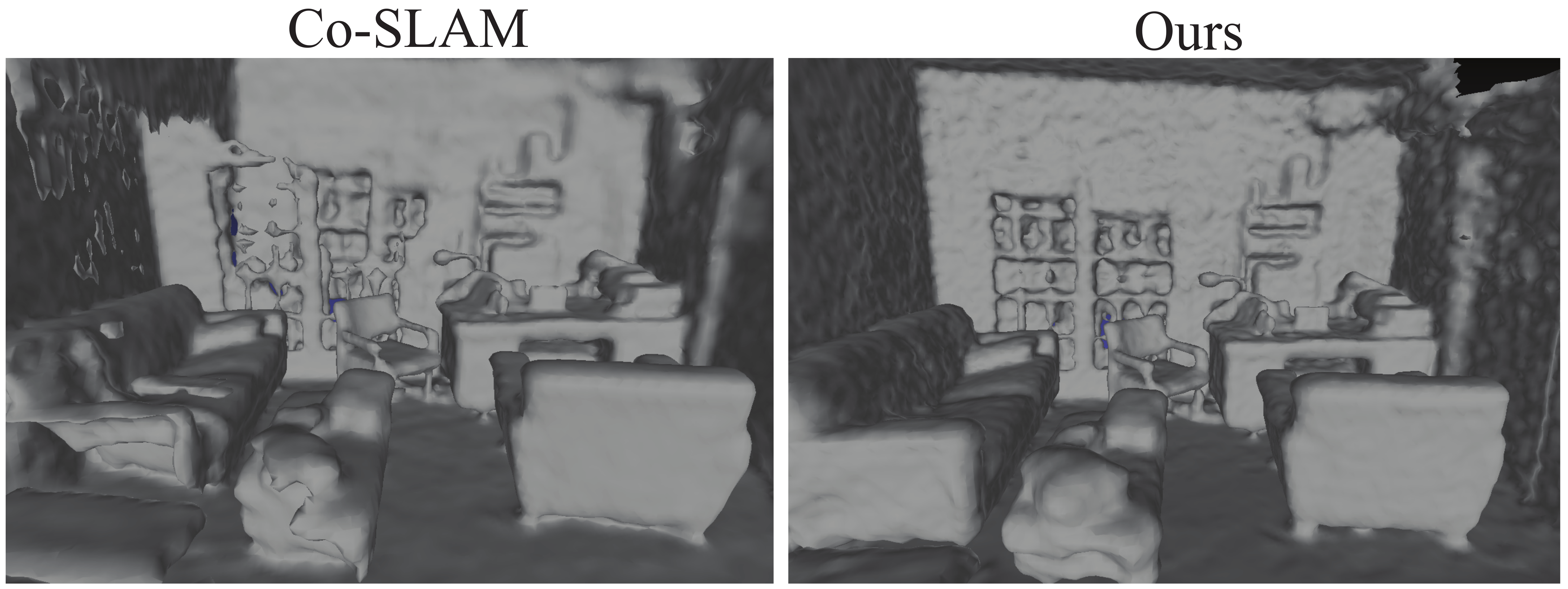}
\caption{\label{fig:scannet}Visual comparison in reconstruction on ScanNet.}
\vspace{-0.2in}
\end{figure}

\noindent\textbf{Implementation Details. }
For query sampling, we sample \( N=43 \) queries per ray, including 32 uniformly sampled and 11 near-surface sampled. We use \( B=128 \) codes for vector quantization and a \( 256^3 \) TSDF resolution with a truncated threshold \( t_r = 10 \) voxel size near surfaces. Following DP Prior~\cite{Hu2023LNI-ADFP}, we incrementally fuse a TSDF using coarsely estimated camera poses. Rays are sampled for volume rendering, and depth fusing is redone with refined poses for the next frame. Loss parameters are set as \( t=0.1, \alpha=0.02, \beta=0.06, \gamma=0.0001, \zeta=200, \eta=2 \).


\begin{figure*}[tb]
  \centering
   \includegraphics[width=0.8\linewidth]{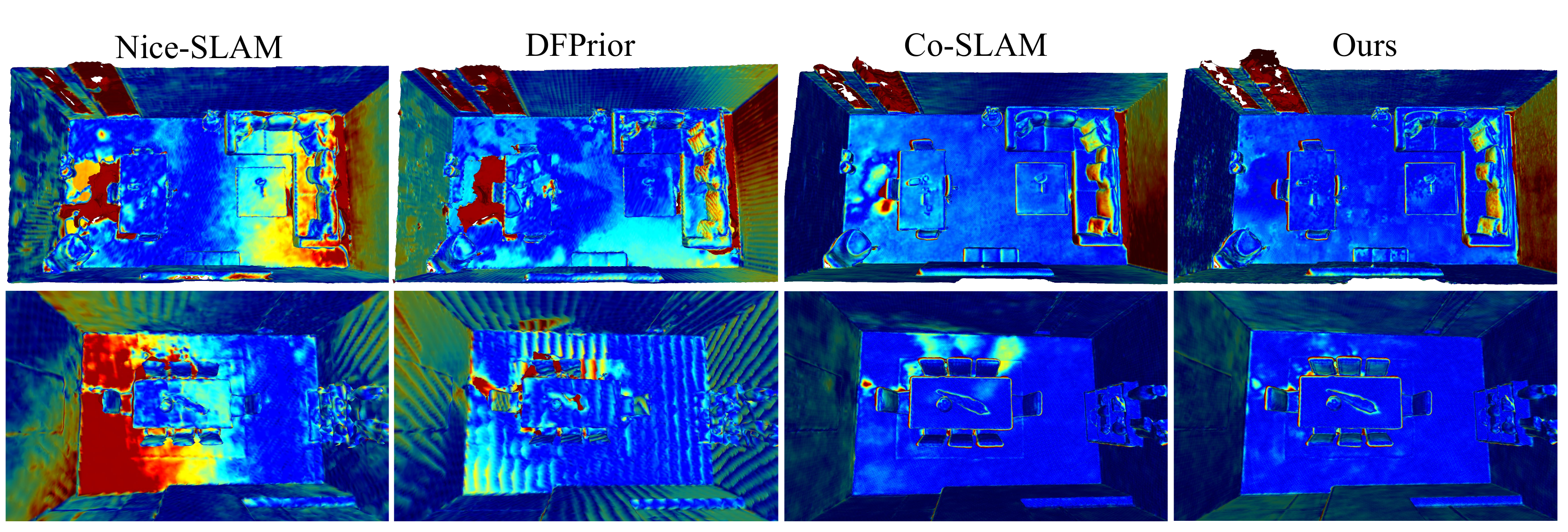}
   \caption{Visual comparisons on Replica.}
   \label{fig:replicacomp}
\end{figure*}

\section{Experiments and Analysis}
\noindent\textbf{Datasets.} We evaluate our method on real-world indoor scenes from 4 datasets and 8 synthetic Replica~\cite{replica19arxiv} scenes following Co-SLAM. Additionally, we assess reconstruction quality on 7 noisy scenes from SyntheticRGBD~\cite{rajpal2023high} and compare our reconstruction and camera tracking accuracy to SOTAs on 6 scenes from NICE-SLAM~\cite{Zhu2021NICESLAM} with BundleFusion ground truth poses. Camera tracking is also reported on 3 scenes from TUM RGB-D~\cite{sturm12iros}.


\begin{figure*}[tb]
  \centering
   \includegraphics[width=0.8\linewidth]{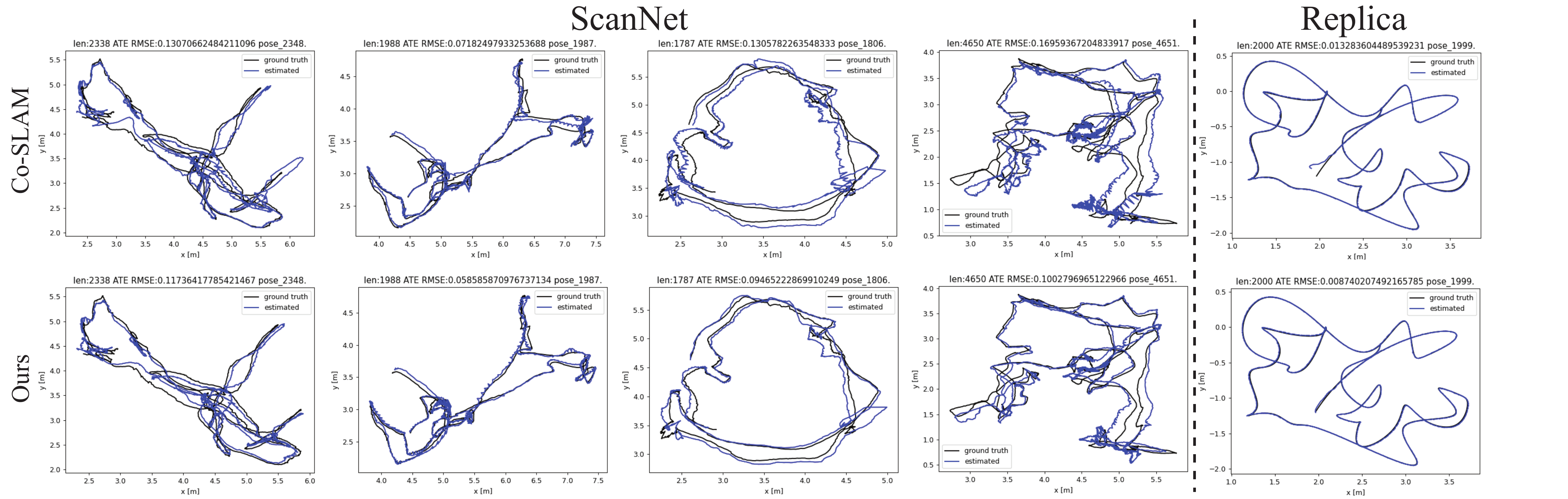}
   \caption{Visual comparisons in camera tracking on ScanNet and Replica.}
   \label{fig:tracking}
   \vspace{-0.12in}
\end{figure*}

\noindent\textbf{Metrics. }We adopt Co-SLAM's culling strategy and evaluate reconstruction accuracy using Depth L1 (cm), Accuracy (cm), Completion (cm), and Completion Ratio (\(<5\text{cm} \%\)). For camera tracking, we report ATE RMSE~\cite{sturm12iros} (cm). Our main baselines include iMAP~\cite{sucar2021imap}, NICE-SLAM~\cite{Zhu2021NICESLAM}, NICER-SLAM~\cite{nicerslam}, DF Prior~\cite{Hu2023LNI-ADFP}, Co-SLAM, and Go-Surf~\cite{wang2022go-surf}, ensuring fair comparison with Co-SLAM's mesh culling.


\begin{table}[tb]
\vspace{-0.1in}
\centering
\resizebox{1.03\linewidth}{!}{
\begin{tabular}{c c c c c c c c c c c} 
\toprule
& &room0 & room1 & room2 & office0 & office1 & office2 & office3 & office4 & Avg.\\
\midrule
{\multirow{4}{*}{\rotatebox[origin=c]{90}{iMAP}}}
&Depth L1[cm]$\downarrow$ &5.08 & 3.44 & 5.78 & 3.79 & 3.76 & 3.97 & 5.61 & 5.71 & 4.64\\
&Acc.[cm]$\downarrow$ &4.01 & 3.04 & 3.84 & 3.34 & 2.10 & 4.06 & 4.20 & 4.34 & 3.62\\
&Comp.[cm]$\downarrow$ &5.84 & 4.40 & 5.07 & 3.62 & 3.62 & 4.73 & 5.49 & 6.65 & 4.93\\
&Comp. Ratio$\uparrow$ &78.34 & 85.85 & 79.40 & 83.59 & 88.45 & 79.73 & 73.90 & 74.77 & 80.50\\
\midrule
{\multirow{4}{*}{\rotatebox[origin=c]{90}{NICE}}}
&Depth L1[cm]$\downarrow$ &1.79 & 1.33 & 2.20 & 1.43 & 1.58 & 2.70 & 2.10 & 2.06 & 1.90\\
&Acc.[cm]$\downarrow$ &2.44 & 2.10 & 2.17 & 1.85 & 1.56 & 3.28 & 3.01 & 2.54 & 2.37\\
&Comp.[cm]$\downarrow$ &2.60 & 2.19 & 2.73 & 1.84 & 1.82 & 3.11 & 3.16 & 3.61 & 2.63\\
&Comp. Ratio $\uparrow$ &91.81 & 93.56 & 91.48 & 94.93 & 94.11 & 88.27 & 87.68 & 87.23 & 91.13\\
\midrule
{\multirow{4}{*}{\rotatebox[origin=c]{90}{DF Prior}}}
&Depth L1[cm]$\downarrow$ &1.44 & 1.90 & 2.75 & 1.43 & 2.03 & 7.73 & 4.81 & 1.99 & 3.01\\
&Acc.[cm]$\downarrow$ &2.54 & 2.70 & 2.25 & 2.14 & 2.80 & 3.58 & 3.46 & 2.68 & 2.77\\
&Comp.[cm]$\downarrow$ &2.41 & 2.26 & 2.46 & 1.76 & 1.94 & 2.56 & 2.93 & 3.27 & 2.45\\
&Comp. Ratio $\uparrow$ &93.22 & 94.75 & 93.02 & 96.04 & 94.77 & 91.89 & 90.17 & 88.46 & 92.79\\
\midrule
{\multirow{4}{*}{\rotatebox[origin=c]{90}{Co-SLAM }}}
&Depth L1[cm]$\downarrow$ &1.05 & 0.85 & 2.37 & 1.24 & 1.48 & 1.86 & 1.66 & 1.54 & 1.51\\
&Acc.[cm]$\downarrow$ &\textbf{2.11} & \textbf{1.68 }&\textbf{ 1.99} & 1.57 & \textbf{1.31} & \textbf{2.84 }& \textbf{3.06} & 2.23 & \textbf{2.10}\\
&Comp.[cm]$\downarrow$ &2.02 & 1.81 & 1.96 & \textbf{1.56} & 1.59 & 2.43 & 2.72 & 2.52 & 2.08\\
&Comp. Ratio $\uparrow$ &95.26 & 95.19 & 93.58 & 96.09 & 94.65 & 91.63 & 90.72 & 90.44 & 93.44\\
\midrule
{\multirow{4}{*}{\rotatebox[origin=c]{90}{Ours }}}
&Depth L1[cm]$\downarrow$ &1.09&\textbf{0.69}&2.48&\textbf{1.18}&\textbf{0.99}&\textbf{1.76}&\textbf{1.54}&1.68&\textbf{1.42}\\
&Acc.[cm]$\downarrow$&2.38 & 2.62 & 2.0 &\textbf{ 1.55} & 1.37 & 3.43 & 3.94 & \textbf{2.16 }& 2.43\\
&Comp.[cm]$\downarrow$ &\textbf{1.76} & \textbf{1.77} & \textbf{1.82} & \textbf{1.57} & \textbf{1.39} & \textbf{2.14} & \textbf{2.55} &\textbf{ 2.46} & \textbf{1.93}\\
&Comp. Ratio $\uparrow$ &\textbf{96.39} & \textbf{95.49} & \textbf{94.28 }& \textbf{96.10 }& \textbf{95.4} &\textbf{ 94.07} & \textbf{91.78} & \textbf{91.53 }& \textbf{94.38}\\
\bottomrule
\end{tabular}}
\vspace{-0.1in}
\caption{Numerical comparison in each scene on Replica.}
\label{table:replicafull}
\vspace{-0.05in}
\end{table}

\begin{figure}[tb]
  \centering
   \includegraphics[width=\linewidth]{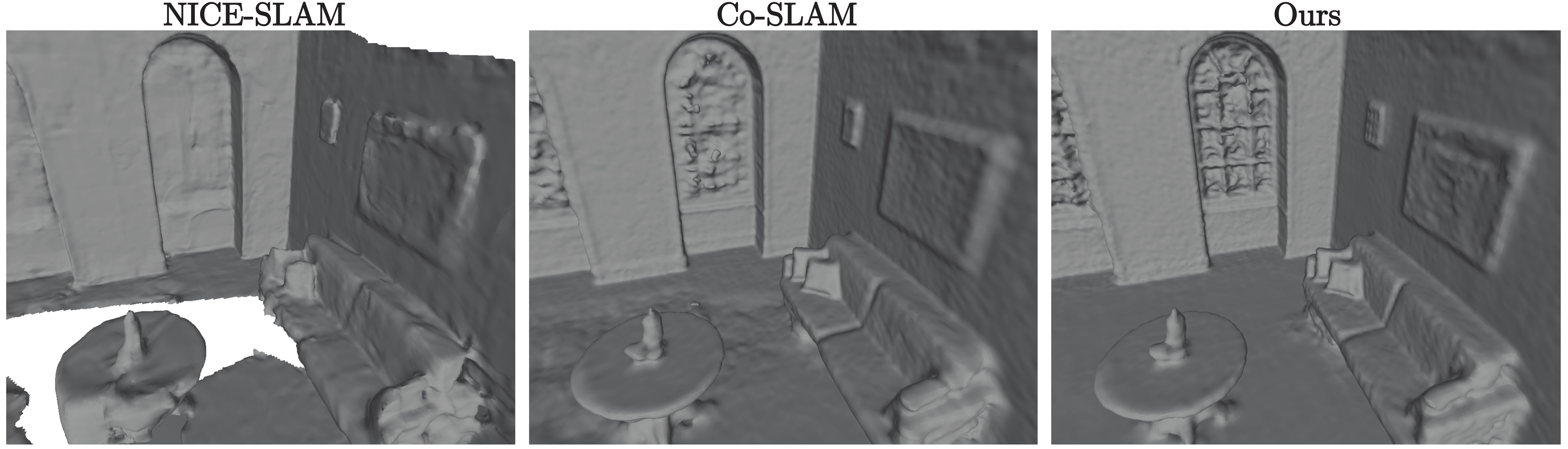}
   \caption{Reconstruction comparisons on SyntheticRGBD.}
   \label{fig:syntheticrgbd}
\end{figure}

\begin{table}[tb]
\centering
\resizebox{0.85\linewidth}{!}{
\begin{tabular}{c c c c} 
\toprule
&Acc.[cm]$\downarrow$   &Comp.[cm]$\downarrow$   &Comp. Ratio $\uparrow$ \\ 
\midrule
NICE-SLAM & \textbf{21.46} & 7.39 & 60.89\\
DF Prior & 22.91 & 8.26 & 52.08\\
Co-SLAM  & 36.89 & 5.75 & 68.46\\
Ours & 39.67 & \textbf{5.09} & \textbf{69.89}\\
\bottomrule
\end{tabular}}
\caption{Reconstruction comparisons on ScanNet.}
\label{table:scannetReconSlam}
\vspace{-0.15in}
\end{table}

\subsection{Evaluations}
\noindent\textbf{Results on Replica. } We evaluate our method on 8 Replica scenes, comparing reconstruction accuracy with iMAP, NICE-SLAM, NICER-SLAM, Co-SLAM, and DF Prior under the same conditions. Tab.~\ref{table:replicafull} shows our method significantly improves surface completion and completion ratios, with visual comparisons in Fig.~\ref{fig:replicacomp}. Our superior reconstruction is due to more accurate camera tracking, as reported in Tab.~\ref{table:atereplica} and visually compared with Co-SLAM in Fig.~\ref{fig:tracking}.


\begin{figure*}[tb]
\vspace{-0.05in}
  \centering
   \includegraphics[width=0.8\linewidth]{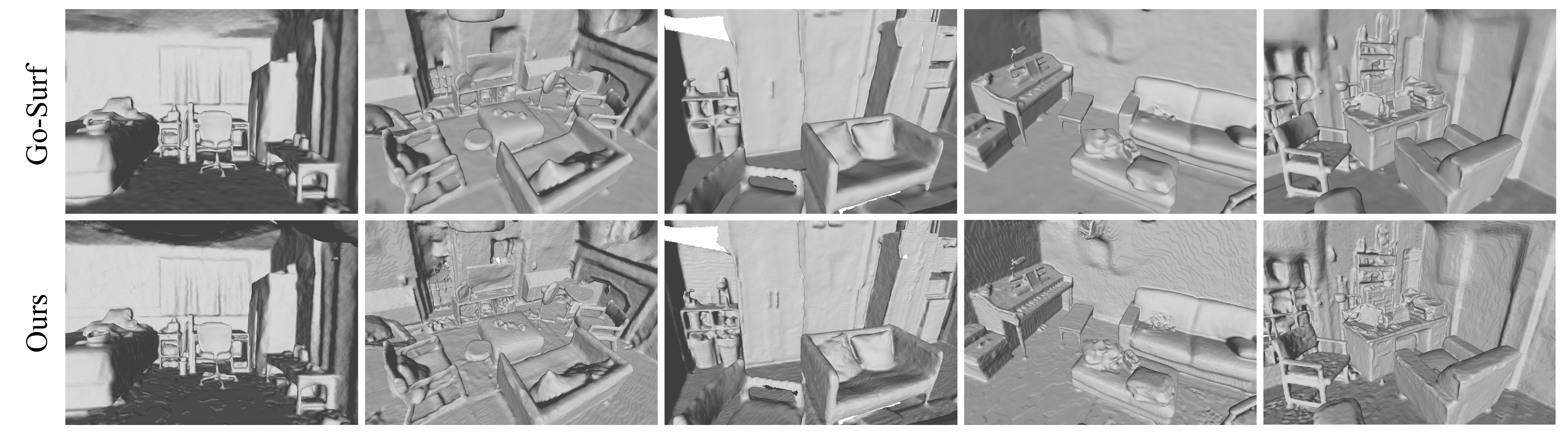}
   \caption{Visual comparisons with Go-Surf on ScanNet.}
   \label{fig:gosurfcomp}
\vspace{-0.12in}
\end{figure*}

\noindent\textbf{Results on SyntheticRGBD. }Tab.~\ref{table:10full} shows numerical comparisons with iMAP, NICE-SLAM, Co-SLAM, and DF Prior on the SyntheticRGBD~\cite{rajpal2023high} dataset. Our method achieves higher accuracy, particularly in completeness and completion ratio metrics. Fig.~\ref{fig:syntheticrgbd} highlights our superior geometric detail, such as window frames and floors in front of sofas. Query quantized neural SLAM reconstructs smoother, more complete surfaces with enhanced detail.

\begin{table}
\vspace{-0.1in}
\centering
\resizebox{0.85\linewidth}{!}{
\begin{tabular}{c c c c} 
\toprule
& fr1/desk (cm) & fr2/xyz (cm) & fr3/office (cm) \\ 
\midrule
iMAP  &4.9 &2.0 & 5.8\\
NICE-SLAM & 2.7 &1.8 &3.0\\
Co-SLAM & 2.7 & 1.9 & \textbf{2.67}\\
Ours & \textbf{2.61} & \textbf{1.7} & 2.70\\
\bottomrule
\end{tabular}}
\caption{ ATE RMSE(cm) in tracking on TUMRGBD.}
\label{table:TUM}
\vspace{-0.05in}
\end{table}

\noindent\textbf{Results on ScanNet. }We evaluate our method on real ScanNet scans. Tab.~\ref{table:scannetReconSlam} shows our method outperforms NICE-SLAM, Co-SLAM, and DF Prior numerically, while Fig.~\ref{fig:scannet} highlights sharper, more compact surfaces. Tab.~\ref{table:ateScannet} and Fig.~\ref{fig:tracking} demonstrate improved camera tracking, particularly on complex real scans, thanks to our quantized queries.

\begin{table}
\centering
\resizebox{\linewidth}{!}{
\begin{tabular}{c c c c c c c c c c c} 
\toprule
& &0000& 0002&0005&0050 &Avg.\\
\midrule
{\multirow{3}{*}{Go-Surf}}
&Acc.[cm]$\downarrow$ &3.18&\textbf{3.48}&14.79&29.36&12.70\\
&Comp.[cm]$\downarrow$ &2.37&2.91&2.04&2.87&2.55\\
&Comp. Ratio [\textless 5cm \%] $\uparrow$&94.04&84.94&92.78&88.31&90.01\\
\hline
{\multirow{3}{*}{Ours }}
&Acc.[cm]$\downarrow$&\textbf{3.17}&4.08&\textbf{13.63}&\textbf{23.45}&\textbf{11.08}\\
&Comp.[cm]$\downarrow$ & \textbf{2.33}&\textbf{2.83}&\textbf{1.97}&\textbf{2.81}&\textbf{2.48}\\
&Comp. Ratio [\textless 5cm \%] $\uparrow$&\textbf{94.3}&\textbf{85.48}&\textbf{94.01}&\textbf{88.38}&\textbf{90.54}\\
\bottomrule
\end{tabular}}
\caption{Numerical comparison in each scene on Scannet.}
\label{table:gosurf}
\vspace{-0.15in}
\end{table}

\noindent\textbf{Results on TUMRGBD. }We follow Co-SLAM to report our tracking performance on TUMRGBD. The numerical comparisons in Tab.~\ref{table:TUM} show that our quantized queries also make networks estimate camera poses more accurately. 

\begin{table}
\centering
\resizebox{0.95\linewidth}{!}{
\begin{tabular}{c c c c c c c c} 
\toprule
Scene ID & 0000 & 0059 & 0106 & 0169 & 0181 & 0207 & Avg.\\ 
\midrule
iMAP & 55.95 &32.06& 17.50 &70.51 &32.10 &11.91 &36.67\\
NICE-SLAM &8.64 &12.25 &\textbf{8.09} &10.28 &\textbf{12.93} &\textbf{5.59} &9.63\\
Co-SLAM  & 7.18 & 12.29 & 9.57 &6.62 &13.43 &7.13& 9.37\\
Ours & \textbf{6.99} & \textbf{9.47} & 8.82& \textbf{6.48} &13.30 &5.86 &\textbf{8.49}\\
\bottomrule
\end{tabular}}
\caption{ATE RMSE(cm) comparisons on ScanNet. }
\label{table:ateScannet}
\vspace{-0.13in}
\end{table}

\noindent\textbf{Application in Multi-View Reconstruction. }We evaluate our quantized queries for multi-view reconstruction using Go-Surf's neural implicit function. Tab.~\ref{table:gosurf} shows our approach consistently outperforms Go-Surf on $4$ ScanNet scenes in Accuracy (cm), Completion (cm), and Completion ratio ($\textless 5cm \%$). Fig.~\ref{fig:gosurfcomp} demonstrates more compact surfaces and enhanced geometric details achieved through better convergence with our quantized queries.

\begin{table}
\vspace{-0.12in}
\centering
\resizebox{0.9\linewidth}{!}{
\begin{tabular}{c c c c c c c c c c} 
\toprule
&rm-0 & rm-1 & rm-2 & off-0 & off-1 & off-2 & off-3 & off-4 & Avg.\\ 
\midrule
NICE & 1.69 &2.04& 1.55& 0.99 & 0.90 & \textbf{1.39} &3.97 &3.08&1.95\\
NICER& 1.36 &1.60 &1.14 &2.12 &3.23 &2.12 & 1.42 & 2.01 & 1.88\\
DF Prior & 1.39 & 1.55 & 2.60 &1.09 &1.23 &1.61& 3.61&1.42&1.81\\
Co-SLAM & 0.72 & 1.32 & 1.27 & 0.62 & 0.52 & 2.07 &1.47 & 0.84 & 1.10\\
Ours & \textbf{0.58} & \textbf{1.16 }& \textbf{0.87} & \textbf{0.52} & \textbf{0.48} & 1.74 &\textbf{1.22} & \textbf{0.73} & \textbf{0.91}\\
\bottomrule
\end{tabular}}
\caption{ATE RMSE(cm) comparisons on Replica.}
\label{table:atereplica}
\end{table}

\subsection{Analysis}
\begin{figure}
  \centering
   \includegraphics[width=0.99\linewidth]{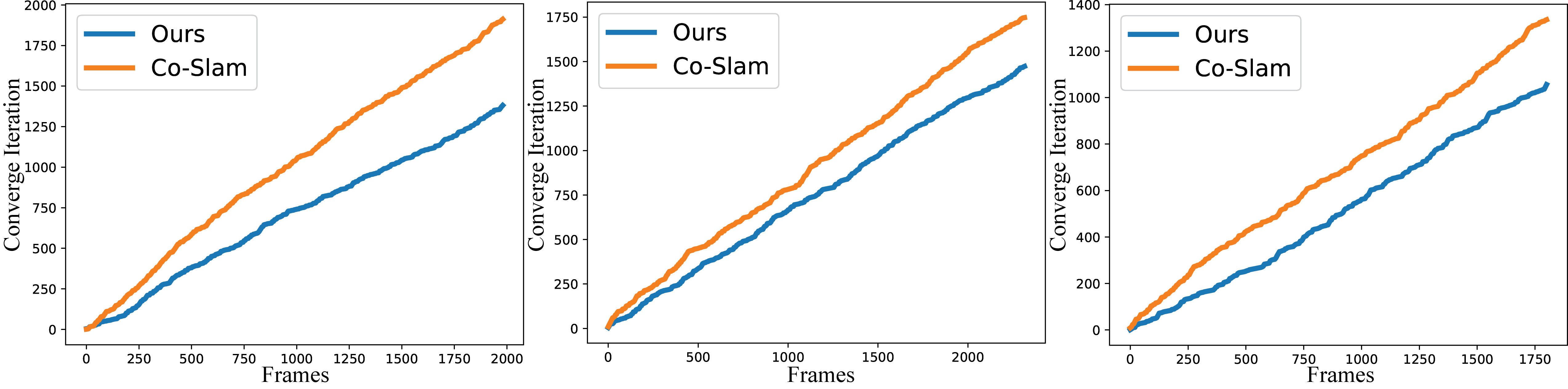}
   \caption{Merits of quantized queries in convergence on scene 0059, 0106, and 0207 from ScanNet.}
   \label{fig:tsne}
   \vspace{-0.12in}
\end{figure}

\noindent\textbf{Why Quantized Queries Work. }For time-sensitive task SLAM, the network can merely get updated in few iterationsf at each frame. Thus, the convergence efficiency is vital to inference accuracy. Our quantized queries significantly reduce the variations of input, which makes neural network always see queries that have been observed at previous frames, leading to fast overfitting on the current frame. 
We record the iteration when our neural network converges at each frame, and visualize the integral of converge iteration at each frame in Fig.~\ref{fig:tsne}. The comparison with Co-SLAM which needs continuous queries show that quantized queries need much fewer iterations than Co-SLAM to converge at a frame. We determine if the optimization converges according to the RGB rendering loss $L_I$ with a threshold of $0.0002$.

\begin{table}
\centering
\resizebox{\linewidth}{!}{
\begin{tabular}{c c c c c c c c c c c} 
\toprule
& &BR& CK &GR &GWR& MA &TG &WR &Avg.\\
\midrule
{\multirow{4}{*}{\rotatebox[origin=c]{90}{iMAP}}}
&Depth L1[cm]$\downarrow$ &24.03 & 63.59 & 26.22 & 21.32 & 61.29 & 29.16 & 81.71 & 47.22\\
&Acc.[cm]$\downarrow$ &10.56 & 25.16 & 13.01 & 11.90 & 29.62 & 12.98 & 24.82 & 18.29\\
&Comp.[cm]$\downarrow$ &11.27 & 31.09 & 19.17 & 20.39 & 49.22 & 21.07 & 32.63 & 26.41\\
&Comp. Ratio $\uparrow$ &46.91 & 12.96 & 21.78 & 20.48 & 10.72 & 19.17 & 13.07 & 20.73\\
\midrule
{\multirow{4}{*}{\rotatebox[origin=c]{90}{NICE}}}
&Depth L1[cm]$\downarrow$ &3.66 & 12.08 & 10.88 & 2.57 & 1.72 & 7.74 & 5.59 & 6.32\\
&Acc.[cm]$\downarrow$ &3.44 & 10.92 & 5.34 & 2.63 & 6.55 & 3.57 & 9.22 & 5.95\\
&Comp.[cm]$\downarrow$ &3.69 & 12.00 & 4.94 & 3.15 & 3.13 & 5.28 & 4.89 & 5.30\\
&Comp. Ratio$\uparrow$ &87.69 & 55.41 & 82.78 & 87.72 & 85.04 & 72.05 & 71.56 & 77.46\\
\midrule
{\multirow{4}{*}{\rotatebox[origin=c]{90}{Co-SLAM}}}
&Depth L1[cm]$\downarrow$ &3.51 & 5.62 & 1.95 & 1.25 & 1.41 & 4.66 & 2.74 & 3.02\\
&Acc.[cm]$\downarrow$ &\textbf{1.97} & \textbf{4.68} & 2.10 & \textbf{1.89} & 1.60 & 3.38 & \textbf{5.03 }& 2.95\\
&Comp.[cm]$\downarrow$ &1.93 &\textbf{ 4.94} & 2.96 & 2.16 & 2.67 & 2.74 & 3.34 & 2.96\\
&Comp. Ratio $\uparrow$ &94.75 & 68.91 & 90.80 & 95.04 & 86.98 & 86.74 & 84.94 & 86.88\\
\midrule
{\multirow{4}{*}{\rotatebox[origin=c]{90}{Ours}}}
&Depth L1[cm]$\downarrow$ &\textbf{3.45}&5.63&\textbf{1.09}&1.46&\textbf{1.28}&\textbf{4.18}&\textbf{2.16}&\textbf{2.75}\\
&Acc.[cm]$\downarrow$ &2.04& 7.16&\textbf{1.83}&2.07& \textbf{1.56}&\textbf{1.63}&5.25&3.07\\
&Comp.[cm]$\downarrow$ &\textbf{1.84}&5.17&\textbf{2.53}&\textbf{2.01}&\textbf{2.66} &\textbf{2.61}&\textbf{3.01}&\textbf{2.83}\\
&Comp. Ratio  $\uparrow$ &\textbf{95.88}&\textbf{69.10}&\textbf{92.44}&\textbf{95.77}& \textbf{88.01}&\textbf{87.75}&\textbf{88.14}&\textbf{88.16}\\
\bottomrule
\end{tabular}}
\vspace{-0.05in}
\caption{Numerical comparison in each scene on Synthetic.}
\label{table:10full}
\vspace{-0.05in}
\end{table}

\begin{figure}[tb]
  \centering
   \includegraphics[width=0.9\linewidth]{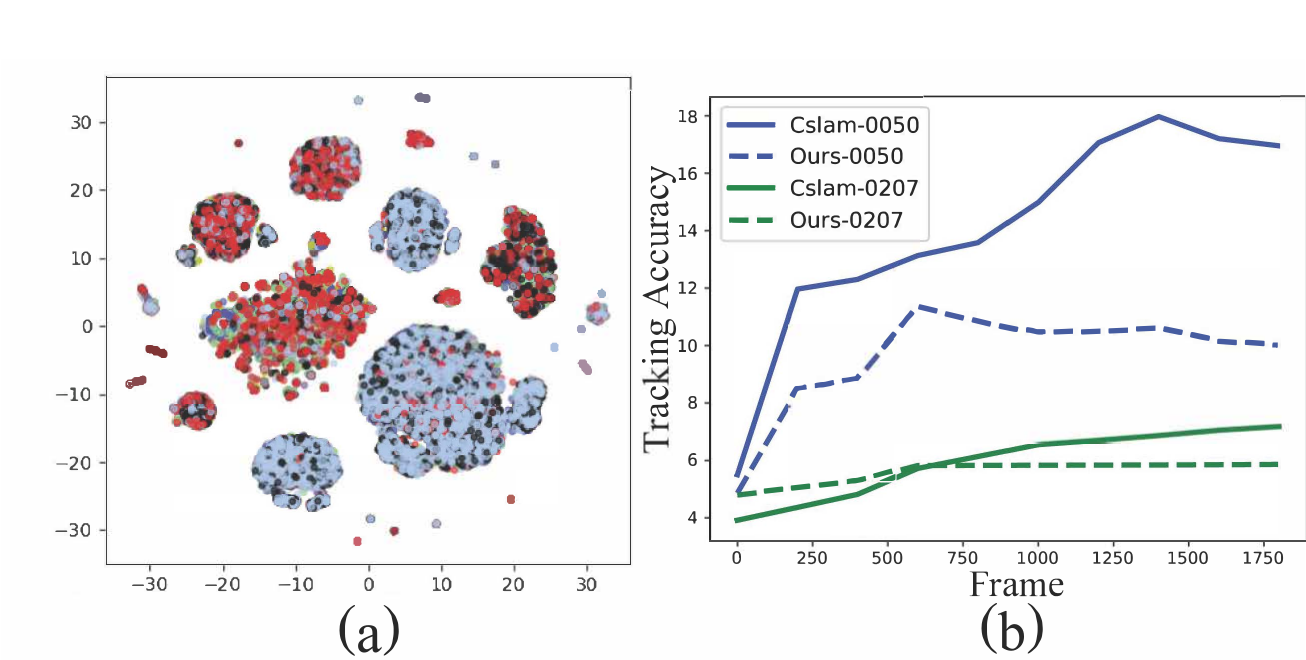}
   \vspace{-0.1in}
   \caption{(a) Codebook visualization with TSNE (Color indicates segmentation labels. Sofa:Red, Wall:Blue.). (b) Comparisons of tracking errors ATE (the lower the better) with Co-SLAM during optimization.
   }
   \label{fig:iterationintegral}
   \vspace{-0.12in}
\end{figure}

Fig.~\ref{fig:iterationintegral} (b) shows the merits of quantized codes in camera tracking. We compare ours with Co-SLAM in terms of errors in different iterations. We see that tracking errors is relatively stable and does not get larger as Co-SLAM once quantized codes converge after about $500$ frames.

\begin{figure}[tb]
    \vspace{-0.06in}
  \centering
   \includegraphics[width=\linewidth]{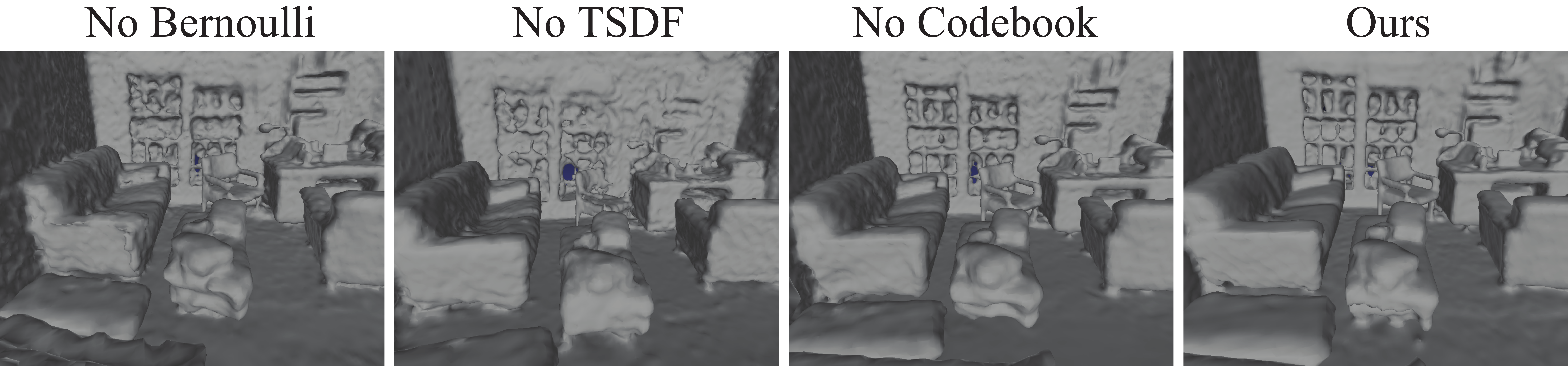}
   \caption{Visual comparisons in ablation studies.}
   \label{fig:ab}
\vspace{-0.16in}
\end{figure}

\begin{table}[tb]
\centering
\resizebox{\linewidth}{!}{
\begin{tabular}{c c c c c c c c} 
\toprule
 & 0050 & 0059 & 0106 & 0207 & Avg.\\ 
\midrule
 w/o Gridcor & 11.34 & 9.96 & 9.01 &6.29&9.15\\
 w/o Codebook &14.76 & 11.24& 9.37 &6.58 &10.49\\
 w/o TSDF& 13.16 &10.53&9.34&6.46&9.87\\
 w/o TSDF1& 11.19&9.87&9.14 &6.17& 9.09 \\
 \midrule
 w/o tanh &14.23&11.39&9.41&6.89 &10.48\\
 w/o Bernoulli & 12.78&10.46&9.17&6.80&9.80\\
 \midrule
 64 codes & 15.15 & 11.50&9.18&6.30&10.53\\
 256 codes & 14.98&10.48&9.37&6.21&10.26\\
 \midrule
 $L_I$&139.00&117.31&201.85&137.53&148.9\\
 $L_I + L_D$&101.76&102.71&221.87&107.22&133.39\\
 $L_I + L_D + L_{g} - L_{e}$&83.97&99.83&84.66 &74.71&85.79\\
 $L_I + L_D + L_{g} - L_{e} + L_{s'}$&13.76&11.01&\textbf{8.78}&6.13&9.92\\
\midrule
Full Model & \textbf{10.02} & \textbf{9.47} & 8.82 & \textbf{5.86}&\textbf{8.54}\\
\bottomrule
\end{tabular}}
\caption{Abalation study on 4 scenes on ScanNet. ATE RMSE(cm) comparisons in tracking.}
\label{table:ablation}
\vspace{-0.12in}
\end{table}

\begin{figure}[tb]
\vspace{-0.13in}
\centering
   \includegraphics[width=0.88\linewidth]{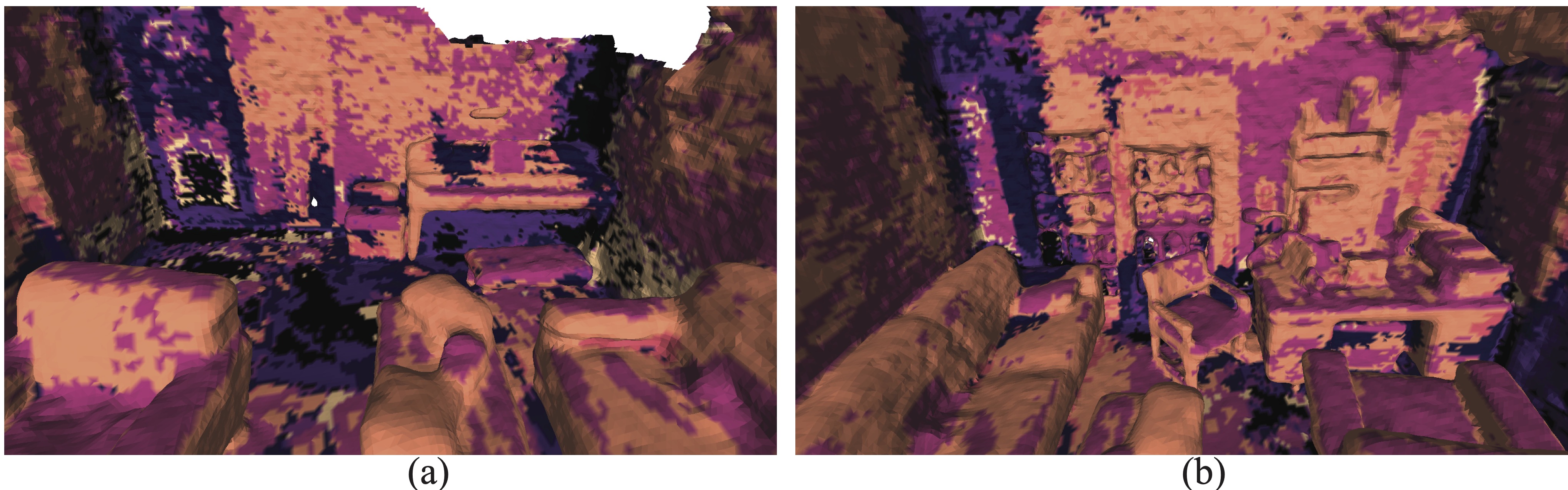}
   \caption{Code ID at vertices on the reconstructed mesh.}
   \label{fig:idshow}
\vspace{-0.1in}
\end{figure}

\noindent\textbf{Code Distribution. }For each vertex on the reconstructed meshes, we query its ID of codes and visualize the ID of codes as color on meshes in Fig.~\ref{fig:idshow}.
We can see that different codes can be used to generate different geometry, and one code can be used to generate similar structures. We also visualize the nearest codes at all vertices with TSNE in Fig.~\ref{fig:iterationintegral} (a) and colorize these codes using the GT segmentation labels of vertices. We can see that codes show some patterns, and several codes which group together may generate the same semantic object like sofa (red) and wall (blue). 

\subsection{Ablation Studies}
\noindent\textbf{Merits of Quantization. }We report merits of quantization on queries in Tab.~\ref{table:ablation}. 

The degenerated results with either continuous coordinate ``w/o Gridcor'', or continuous geometry features ``w/o Codebook'', or no geometry prior ``w/o TSDF'', or continuous geometry prior ``w/o TSDF1'' show the advantages of quantized queries. Fig.~\ref{fig:ab} shows the visual comparisons. We can see that the reconstruction degenerates without the geometry prior TSDF, and we can not estimate accurate zero level set if no codebook is used.

\noindent\textbf{Code Initialization. }We conduct experiments to highlight the importance of codebook initialization. We try to use other distribution like uniform distribution to replace the Bernoulli distribution in the code initialization. The result ``w/o Bernoulli'' in Tab.~\ref{table:ablation} and Fig.~\ref{fig:ab} (a) shows that Bernoulli distribution can significantly stabilize the optimization by constraining the optimization space with uncertainties since very beginning. This initialization is vital to our method.

\noindent\textbf{Effectiveness of Loss Terms. }We justify the effectiveness of each loss term in Tab.~\ref{table:ablation}. We incrementally add one loss term each time. We can see that each loss can improve the tracking performance.

\noindent\textbf{Effect of Code Number. }We also explore the effect of code number in Tab.~\ref{table:ablation}. We try different numbers including $B=\{64,256\}$. We see that either fewer or more codes degenerate the performance. This is because too few codes are not enough to represent diverse geometries while too many codes make it hard to learn patterns well as codes in the overfitting on a scene.

\noindent\textbf{Effect of TSDF Augmentation. }Using singed distances interpolated from TSDF fusion as a geometry prior also improves reconstruction accuracy. We report results ``w/o tanh'' without signed distance interpolations in Tab.~\ref{table:ablation}. The degenerated results justify its effectiveness and the importance of making the input different to the output.

\section{Conclusion}
We present query quantized neural SLAM for joint camera pose estimation and scene reconstruction. By quantizing queries including coordinates, positional encodings, geometry features, or priors, we reduce query variations, enabling faster neural network convergence per frame. Our novel initialization, losses, and augmentations stabilize optimization, making quantized coordinates effective for neural SLAM. Extensive evaluations on widely used benchmarks show our method outperforms existing approaches in both camera tracking and reconstruction accuracy.

{
    \small
    \bibliography{papers}

\begin{thebibliography}{70}
\providecommand{\natexlab}[1]{#1}

\bibitem[{Atzmon and Lipman(2021)}]{atzmon2020sald}
Atzmon, M.; and Lipman, y. 2021.
\newblock {SALD:} Sign Agnostic Learning with Derivatives.
\newblock In \emph{International Conference on Learning Representations}.

\bibitem[{Azinovi\'c et~al.(2022)Azinovi\'c, Martin-Brualla, Goldman, Nie{\ss}ner, and Thies}]{Azinovic_2022_CVPR}
Azinovi\'c, D.; Martin-Brualla, R.; Goldman, D.~B.; Nie{\ss}ner, M.; and Thies, J. 2022.
\newblock Neural RGB-D Surface Reconstruction.
\newblock In \emph{IEEE Conference on Computer Vision and Pattern Recognition}, 6290--6301.

\bibitem[{Chen, Liu, and Han(2022)}]{chaompi2022}
Chen, C.; Liu, Y.-S.; and Han, Z. 2022.
\newblock Latent Partition Implicit with Surface Codes for 3D Representation.
\newblock In \emph{European Conference on Computer Vision}.

\bibitem[{Chen, Liu, and Han(2023{\natexlab{a}})}]{chao2023gridpull}
Chen, C.; Liu, Y.-S.; and Han, Z. 2023{\natexlab{a}}.
\newblock GridPull: Towards Scalability in Learning Implicit Representations from 3D Point Clouds.
\newblock In \emph{IEEE International Conference on Computer Vision}.

\bibitem[{Chen, Liu, and Han(2023{\natexlab{b}})}]{ChaoSparse}
Chen, C.; Liu, Y.-S.; and Han, Z. 2023{\natexlab{b}}.
\newblock Unsupervised Inference of Signed Distance Functions from Single Sparse Point Clouds without Learning Priors.
\newblock In \emph{Proceedings of the IEEE/CVF Conference on Computer Vsion and Pattern Recognition}.

\bibitem[{Chen, Liu, and Han(2024)}]{localn2nm2024}
Chen, C.; Liu, Y.-S.; and Han, Z. 2024.
\newblock Inferring Neural Signed Distance Functions by Overfitting on Single Noisy Point Clouds through Finetuning Data-Driven based Priors.
\newblock In \emph{Advances in Neural Information Processing Systems}.

\bibitem[{Corona-Figueroa et~al.(2023)Corona-Figueroa, Bond-Taylor, Bhowmik, Gaus, Breckon, Shum, and Willcocks}]{corona2023unaligned}
Corona-Figueroa, A.; Bond-Taylor, S.; Bhowmik, N.; Gaus, Y. F.~A.; Breckon, T.~P.; Shum, H.~P.; and Willcocks, C.~G. 2023.
\newblock Unaligned 2D to 3D Translation with Conditional Vector-Quantized Code Diffusion using Transformers.
\newblock In \emph{IEEE/CVF International Conference on Computer Vision}, 14585--14594.

\bibitem[{Dupont et~al.(2022)Dupont, Loya, Alizadeh, Goli{\'n}ski, Teh, and Doucet}]{dupont2022coin++}
Dupont, E.; Loya, H.; Alizadeh, M.; Goli{\'n}ski, A.; Teh, Y.~W.; and Doucet, A. 2022.
\newblock COIN++: Neural compression across modalities.
\newblock \emph{arXiv preprint arXiv:2201.12904}.

\bibitem[{Fei et~al.(2022)Fei, Yang, Chen, and Ma}]{fei2022vq}
Fei, B.; Yang, W.; Chen, W.-M.; and Ma, L. 2022.
\newblock VQ-DcTr: Vector-quantized autoencoder with dual-channel transformer points splitting for 3D point cloud completion.
\newblock In \emph{30th ACM international conference on multimedia}, 4769--4778.

\bibitem[{Fu et~al.(2022)Fu, Xu, Ong, and Tao}]{GEOnEUS2022}
Fu, Q.; Xu, Q.; Ong, Y.-S.; and Tao, W. 2022.
\newblock {Geo-Neus}: Geometry-Consistent Neural Implicit Surfaces Learning for Multi-view Reconstruction.
\newblock In \emph{Advances in Neural Information Processing Systems}.

\bibitem[{Gordon et~al.(2023)Gordon, Chng, MacDonald, and Lucey}]{gordon2023quantizing}
Gordon, C.; Chng, S.-F.; MacDonald, L.; and Lucey, S. 2023.
\newblock On Quantizing Implicit Neural Representations.
\newblock In \emph{IEEE/CVF Winter Conference on Applications of Computer Vision}, 341--350.

\bibitem[{Gu et~al.(2022)Gu, Chen, Bao, Wen, Zhang, Chen, Yuan, and Guo}]{gu2022vector}
Gu, S.; Chen, D.; Bao, J.; Wen, F.; Zhang, B.; Chen, D.; Yuan, L.; and Guo, B. 2022.
\newblock Vector quantized diffusion model for text-to-image synthesis.
\newblock In \emph{Proceedings of the IEEE/CVF Conference on Computer Vision and Pattern Recognition}, 10696--10706.

\bibitem[{Guo et~al.(2022)Guo, Peng, Lin, Wang, Zhang, Bao, and Zhou}]{guo2022manhattan}
Guo, H.; Peng, S.; Lin, H.; Wang, Q.; Zhang, G.; Bao, H.; and Zhou, X. 2022.
\newblock Neural 3D Scene Reconstruction with the Manhattan-world Assumption.
\newblock In \emph{{IEEE} Conference on Computer Vision and Pattern Recognition}.

\bibitem[{Haghighi et~al.(2023)Haghighi, Kumar, Thiran, and Gool}]{haghighi2023neural}
Haghighi, Y.; Kumar, S.; Thiran, J.-P.; and Gool, L.~V. 2023.
\newblock Neural Implicit Dense Semantic SLAM.
\newblock arXiv:2304.14560.

\bibitem[{Hu and Han(2023)}]{Hu2023LNI-ADFP}
Hu, P.; and Han, Z. 2023.
\newblock Learning Neural Implicit through Volume Rendering with Attentive Depth Fusion Priors.
\newblock In \emph{Advances in Neural Information Processing Systems (NeurIPS)}.

\bibitem[{Huang et~al.(2024{\natexlab{a}})Huang, Yu, Chen, Geiger, and Gao}]{Huang_2024}
Huang, B.; Yu, Z.; Chen, A.; Geiger, A.; and Gao, S. 2024{\natexlab{a}}.
\newblock 2D Gaussian Splatting for Geometrically Accurate Radiance Fields.
\newblock In \emph{Special Interest Group on Computer Graphics and Interactive Techniques Conference Conference Papers ’24}, SIGGRAPH ’24. ACM.

\bibitem[{Huang et~al.(2024{\natexlab{b}})Huang, Li, Hui, and Yeung}]{hhuang2024photoslam}
Huang, H.; Li, L.; Hui, C.; and Yeung, S.-K. 2024{\natexlab{b}}.
\newblock Photo-SLAM: Real-time Simultaneous Localization and Photorealistic Mapping for Monocular, Stereo, and RGB-D Cameras.
\newblock In \emph{IEEE/CVF Conference on Computer Vision and Pattern Recognition}.

\bibitem[{Jiang, Hua, and Han(2023)}]{sijia2023quantized}
Jiang, S.; Hua, J.; and Han, Z. 2023.
\newblock Coordinate Quantized Neural Implicit Representations for Multi-view 3D Reconstruction.
\newblock In \emph{{IEEE} International Conference on Computer Vision}.

\bibitem[{Keetha et~al.(2024)Keetha, Karhade, Jatavallabhula, Yang, Scherer, Ramanan, and Luiten}]{keetha2024splatam}
Keetha, N.; Karhade, J.; Jatavallabhula, K.~M.; Yang, G.; Scherer, S.; Ramanan, D.; and Luiten, J. 2024.
\newblock SplaTAM: Splat, Track \& Map 3D Gaussians for Dense RGB-D SLAM.
\newblock In \emph{IEEE/CVF Conference on Computer Vision and Pattern Recognition}.

\bibitem[{Koestler et~al.(2022)Koestler, Yang, Zeller, and Cremers}]{koestler2022tandem}
Koestler, L.; Yang, N.; Zeller, N.; and Cremers, D. 2022.
\newblock Tandem: Tracking and dense mapping in real-time using deep multi-view stereo.
\newblock In \emph{Conference on Robot Learning}, 34--45. PMLR.

\bibitem[{Kong et~al.(2023)Kong, Liu, Taher, and Davison}]{kong2023vmap}
Kong, X.; Liu, S.; Taher, M.; and Davison, A.~J. 2023.
\newblock vMAP: Vectorised Object Mapping for Neural Field SLAM.
\newblock \emph{arXiv preprint arXiv:2302.01838}.

\bibitem[{Laurentini(1994)}]{273735visualhull}
Laurentini, A. 1994.
\newblock The visual hull concept for silhouette-based image understanding.
\newblock \emph{IEEE Transactions on Pattern Analysis and Machine Intelligence}, 16(2): 150--162.

\bibitem[{Lee et~al.(2023)Lee, Park, Son, Ryu, and Chae}]{lee2023fastsurf}
Lee, S.; Park, G.; Son, H.; Ryu, J.; and Chae, H.~J. 2023.
\newblock FastSurf: Fast Neural RGB-D Surface Reconstruction using Per-Frame Intrinsic Refinement and TSDF Fusion Prior Learning.
\newblock \emph{arXiv preprint arXiv:2303.04508}.

\bibitem[{Li et~al.(2023{\natexlab{a}})Li, Dou, Chen, Ni, Sun, Liu, and Wang}]{li2023generalized}
Li, Y.; Dou, Y.; Chen, X.; Ni, B.; Sun, Y.; Liu, Y.; and Wang, F. 2023{\natexlab{a}}.
\newblock Generalized Deep 3D Shape Prior via Part-Discretized Diffusion Process.
\newblock In \emph{Proceedings of the IEEE/CVF Conference on Computer Vision and Pattern Recognition}, 16784--16794.

\bibitem[{Li et~al.(2023{\natexlab{b}})Li, M\"uller, Evans, Taylor, Unberath, Liu, and Lin}]{li2023neuralangelo}
Li, Z.; M\"uller, T.; Evans, A.; Taylor, R.~H.; Unberath, M.; Liu, M.-Y.; and Lin, C.-H. 2023{\natexlab{b}}.
\newblock Neuralangelo: High-Fidelity Neural Surface Reconstruction.
\newblock In \emph{IEEE Conference on Computer Vision and Pattern Recognition}.

\bibitem[{Liu et~al.(2021)Liu, Guo, Pan, Wang, Tong, and Liu}]{Liu2021MLS}
Liu, S.-L.; Guo, H.-X.; Pan, H.; Wang, P.; Tong, X.; and Liu, Y. 2021.
\newblock Deep Implicit Moving Least-Squares Functions for {3D} Reconstruction.
\newblock In \emph{IEEE Conference on Computer Vision and Pattern Recognition}.

\bibitem[{Lorensen and Cline(1987)}]{Lorensen87marchingcubes}
Lorensen, W.~E.; and Cline, H.~E. 1987.
\newblock Marching cubes: A high resolution {3D} surface construction algorithm.
\newblock \emph{Computer Graphics}, 21(4): 163--169.

\bibitem[{Ma et~al.(2023)Ma, Zhou, Liu, and Han}]{Baoruicvpr2023}
Ma, B.; Zhou, J.; Liu, Y.-S.; and Han, Z. 2023.
\newblock Towards Better Gradient Consistency for Neural Signed Distance Functions via Level Set Alignment.
\newblock In \emph{IEEE/CVF Conference on Computer Vsion and Pattern Recognition}.

\bibitem[{Matsuki et~al.(2024)Matsuki, Murai, Kelly, and Davison}]{MatsukiCVPR2024}
Matsuki, H.; Murai, R.; Kelly, P. H.~J.; and Davison, A.~J. 2024.
\newblock {G}aussian {S}platting {SLAM}.

\bibitem[{Mildenhall et~al.(2020)Mildenhall, Srinivasan, Tancik, Barron, Ramamoorthi, and Ng}]{mildenhall2020nerf}
Mildenhall, B.; Srinivasan, P.~P.; Tancik, M.; Barron, J.~T.; Ramamoorthi, R.; and Ng, R. 2020.
\newblock {NeRF}: Representing Scenes as Neural Radiance Fields for View Synthesis.
\newblock In \emph{European Conference on Computer Vision}.

\bibitem[{M\"uller et~al.(2022)M\"uller, Evans, Schied, and Keller}]{mueller2022instant}
M\"uller, T.; Evans, A.; Schied, C.; and Keller, A. 2022.
\newblock Instant Neural Graphics Primitives with a Multiresolution Hash Encoding.
\newblock \emph{arXiv:2201.05989}.

\bibitem[{M{\"u}ller et~al.(2019)M{\"u}ller, McWilliams, Rousselle, Gross, and Nov{\'a}k}]{muller2019neural}
M{\"u}ller, T.; McWilliams, B.; Rousselle, F.; Gross, M.; and Nov{\'a}k, J. 2019.
\newblock Neural importance sampling.
\newblock \emph{ACM Transactions on Graphics (ToG)}, 38(5): 1--19.

\bibitem[{Niemeyer et~al.(2020)Niemeyer, Mescheder, Oechsle, and Geiger}]{DVRcvpr}
Niemeyer, M.; Mescheder, L.; Oechsle, M.; and Geiger, A. 2020.
\newblock Differentiable Volumetric Rendering: Learning Implicit 3D Representations without 3D Supervision.
\newblock In \emph{IEEE Conference on Computer Vision and Pattern Recognition}.

\bibitem[{Noda et~al.(2024)Noda, Chen, Zhang, Liu, Liu, and Han}]{multigrid}
Noda, T.; Chen, C.; Zhang, W.; Liu, X.; Liu, Y.-S.; and Han, Z. 2024.
\newblock MultiPull: Detailing Signed Distance Functions by Pulling Multi-Level Queries at Multi-Step.
\newblock In \emph{Advances in Neural Information Processing Systems}.

\bibitem[{Oechsle, Peng, and Geiger(2021)}]{Oechsle2021ICCV}
Oechsle, M.; Peng, S.; and Geiger, A. 2021.
\newblock {UNISURF}: Unifying Neural Implicit Surfaces and Radiance Fields for Multi-View Reconstruction.
\newblock In \emph{International Conference on Computer Vision}.

\bibitem[{Oord, Vinyals, and Kavukcuoglu(2017)}]{oord2017neural}
Oord, A. v.~d.; Vinyals, O.; and Kavukcuoglu, K. 2017.
\newblock Neural discrete representation learning.
\newblock \emph{arXiv preprint arXiv:1711.00937}.

\bibitem[{Park et~al.(2021)Park, Sinha, Barron, Bouaziz, Goldman, Seitz, and Martin-Brualla}]{park2021nerfies}
Park, K.; Sinha, U.; Barron, J.~T.; Bouaziz, S.; Goldman, D.~B.; Seitz, S.~M.; and Martin-Brualla, R. 2021.
\newblock Nerfies: Deformable Neural Radiance Fields.
\newblock \emph{ICCV}.

\bibitem[{Peng et~al.(2020)Peng, Niemeyer, Mescheder, Pollefeys, and Geiger}]{Peng2020ECCV}
Peng, S.; Niemeyer, M.; Mescheder, L.; Pollefeys, M.; and Geiger, A. 2020.
\newblock Convolutional occupancy networks.
\newblock In \emph{Computer Vision--ECCV 2020: 16th European Conference, Glasgow, UK, August 23--28, 2020, Proceedings, Part III 16}, 523--540. Springer.

\bibitem[{Rajpal et~al.(2023)Rajpal, Cheema, Illgner-Fehns, Slusallek, and Jaiswal}]{rajpal2023high}
Rajpal, A.; Cheema, N.; Illgner-Fehns, K.; Slusallek, P.; and Jaiswal, S. 2023.
\newblock High-Resolution Synthetic RGB-D Datasets for Monocular Depth Estimation.
\newblock In \emph{CVPR}, 1188--1198.

\bibitem[{Rosu and Behnke(2023)}]{rosu2023permutosdf}
Rosu, R.~A.; and Behnke, S. 2023.
\newblock PermutoSDF: Fast Multi-View Reconstruction with Implicit Surfaces using Permutohedral Lattices.
\newblock In \emph{IEEE/CVF Conference on Computer Vision and Pattern Recognition (CVPR)}.

\bibitem[{Sandström et~al.(2023)Sandström, Ta, Gool, and Oswald}]{uncleslam2023}
Sandström, E.; Ta, K.; Gool, L.~V.; and Oswald, M.~R. 2023.
\newblock Uncle-{SLAM}: Uncertainty Learning for Dense Neural {SLAM}.
\newblock In \emph{International Conference on Computer Vision Workshops (ICCVW)}.

\bibitem[{Sch\"{o}nberger and Frahm(2016)}]{schoenberger2016sfm}
Sch\"{o}nberger, J.~L.; and Frahm, J.-M. 2016.
\newblock Structure-from-Motion Revisited.
\newblock In \emph{{IEEE} Conference on Computer Vision and Pattern Recognition}.

\bibitem[{Sch\"{o}nberger et~al.(2016)Sch\"{o}nberger, Zheng, Pollefeys, and Frahm}]{schoenberger2016mvs}
Sch\"{o}nberger, J.~L.; Zheng, E.; Pollefeys, M.; and Frahm, J.-M. 2016.
\newblock Pixelwise View Selection for Unstructured Multi-View Stereo.
\newblock In \emph{European Conference on Computer Vision}.

\bibitem[{Stier et~al.(2023)Stier, Ranjan, Colburn, Yan, Yang, Ma, and Angles}]{stier2023finerecon}
Stier, N.; Ranjan, A.; Colburn, A.; Yan, Y.; Yang, L.; Ma, F.; and Angles, B. 2023.
\newblock {FineRecon}: Depth-aware Feed-forward Network for Detailed 3D Reconstruction.
\newblock \emph{arXiv preprint}.

\bibitem[{Straub et~al.(2019)Straub, Whelan, Ma, Chen, Wijmans, Green, Engel, Mur-Artal, Ren, Verma, Clarkson, Yan, Budge, Yan, Pan, Yon, Zou, Leon, Carter, Briales, Gillingham, Mueggler, Pesqueira, Savva, Batra, Strasdat, Nardi, Goesele, Lovegrove, and Newcombe}]{replica19arxiv}
Straub, J.; Whelan, T.; Ma, L.; Chen, Y.; Wijmans, E.; Green, S.; Engel, J.~J.; Mur-Artal, R.; Ren, C.; Verma, S.; Clarkson, A.; Yan, M.; Budge, B.; Yan, Y.; Pan, X.; Yon, J.; Zou, Y.; Leon, K.; Carter, N.; Briales, J.; Gillingham, T.; Mueggler, E.; Pesqueira, L.; Savva, M.; Batra, D.; Strasdat, H.~M.; Nardi, R.~D.; Goesele, M.; Lovegrove, S.; and Newcombe, R. 2019.
\newblock The {R}eplica Dataset: A Digital Replica of Indoor Spaces.
\newblock \emph{arXiv preprint arXiv:1906.05797}.

\bibitem[{Sturm et~al.(2012)Sturm, Engelhard, Endres, Burgard, and Cremers}]{sturm12iros}
Sturm, J.; Engelhard, N.; Endres, F.; Burgard, W.; and Cremers, D. 2012.
\newblock A Benchmark for the Evaluation of RGB-D SLAM Systems.
\newblock In \emph{International Conference on Intelligent Robot Systems (IROS)}.

\bibitem[{Sucar et~al.(2021)Sucar, Liu, Ortiz, and Davison}]{sucar2021imap}
Sucar, E.; Liu, S.; Ortiz, J.; and Davison, A.~J. 2021.
\newblock iMAP: Implicit mapping and positioning in real-time.
\newblock In \emph{IEEE/CVF International Conference on Computer Vision}, 6229--6238.

\bibitem[{Sun et~al.(2021)Sun, Xie, Chen, Zhou, and Bao}]{sun2021neucon}
Sun, J.; Xie, Y.; Chen, L.; Zhou, X.; and Bao, H. 2021.
\newblock {NeuralRecon}: Real-Time Coherent {3D} Reconstruction from Monocular Video.
\newblock \emph{CVPR}.

\bibitem[{Tang et~al.(2021)Tang, Lei, Xu, Ma, Jia, and Zhang}]{tang2021sign}
Tang, J.; Lei, J.; Xu, D.; Ma, F.; Jia, K.; and Zhang, L. 2021.
\newblock {SA-ConvONet}: Sign-Agnostic Optimization of Convolutional Occupancy Networks.
\newblock In \emph{ICCV}.

\bibitem[{Teigen et~al.(2023)Teigen, Park, Stahl, and Mester}]{teigen2023rgb}
Teigen, A.~L.; Park, Y.; Stahl, A.; and Mester, R. 2023.
\newblock RGB-D Mapping and Tracking in a Plenoxel Radiance Field.
\newblock \emph{arXiv preprint arXiv:2307.03404}.

\bibitem[{Wang, Wang, and Agapito(2023)}]{wang2023coslam}
Wang, H.; Wang, J.; and Agapito, L. 2023.
\newblock Co-SLAM: Joint Coordinate and Sparse Parametric Encodings for Neural Real-Time SLAM.
\newblock arXiv:2304.14377.

\bibitem[{Wang, Bleja, and Agapito(2022)}]{wang2022go-surf}
Wang, J.; Bleja, T.; and Agapito, L. 2022.
\newblock GO-Surf: Neural Feature Grid Optimization for Fast, High-Fidelity RGB-D Surface Reconstruction.
\newblock In \emph{International Conference on 3D Vision}.

\bibitem[{Wang et~al.(2022)Wang, Wang, Long, Theobalt, Komura, Liu, and Wang}]{wang2022neuris}
Wang, J.; Wang, P.; Long, X.; Theobalt, C.; Komura, T.; Liu, L.; and Wang, W. 2022.
\newblock {NeuRIS}: Neural Reconstruction of Indoor Scenes Using Normal Priors.
\newblock In \emph{European Conference on Computer Vision}.

\bibitem[{Wang et~al.(2021)Wang, Liu, Liu, Theobalt, Komura, and Wang}]{neuslingjie}
Wang, P.; Liu, L.; Liu, Y.; Theobalt, C.; Komura, T.; and Wang, W. 2021.
\newblock {NeuS}: Learning Neural Implicit Surfaces by Volume Rendering for Multi-view Reconstruction.
\newblock In \emph{Advances in Neural Information Processing Systems}, 27171--27183.

\bibitem[{Wu et~al.(2022)Wu, Lei, Sun, Wang, Chen, Cheng, Lin, and Wu}]{wu2022randomized}
Wu, H.; Lei, C.; Sun, X.; Wang, P.-S.; Chen, Q.; Cheng, K.-T.; Lin, S.; and Wu, Z. 2022.
\newblock Randomized Quantization for Data Agnostic Representation Learning.
\newblock \emph{arXiv preprint arXiv:2212.08663}.

\bibitem[{Xinyang et~al.(2023)Xinyang, Yijin, Yanbin, Hujun, Guofeng, Yinda, and Zhaopeng}]{tofslam}
Xinyang, L.; Yijin, L.; Yanbin, T.; Hujun, B.; Guofeng, Z.; Yinda, Z.; and Zhaopeng, C. 2023.
\newblock Multi-Modal Neural Radiance Field for Monocular Dense SLAM with a Light-Weight ToF Sensor.
\newblock In \emph{International Conference on Computer Vision (ICCV)}.

\bibitem[{Yang et~al.(2023{\natexlab{a}})Yang, Lin, Chen, and Zhou}]{yang2023neural}
Yang, X.; Lin, G.; Chen, Z.; and Zhou, L. 2023{\natexlab{a}}.
\newblock Neural Vector Fields: Implicit Representation by Explicit Learning.
\newblock In \emph{Proceedings of the IEEE/CVF Conference on Computer Vision and Pattern Recognition}, 16727--16738.

\bibitem[{Yang et~al.(2023{\natexlab{b}})Yang, Liu, Yin, Chen, Yu, Fan, and Chen}]{yang2023vq}
Yang, Y.; Liu, W.; Yin, F.; Chen, X.; Yu, G.; Fan, J.; and Chen, T. 2023{\natexlab{b}}.
\newblock VQ-NeRF: Vector Quantization Enhances Implicit Neural Representations.
\newblock \emph{arXiv preprint arXiv:2310.14487}.

\bibitem[{Yao et~al.(2018)Yao, Luo, Li, Fang, and Quan}]{yao2018mvsnet}
Yao, Y.; Luo, Z.; Li, S.; Fang, T.; and Quan, L. 2018.
\newblock MVSNet: Depth Inference for Unstructured Multi-view Stereo.
\newblock \emph{European Conference on Computer Vision}.

\bibitem[{Yariv et~al.(2020)Yariv, Kasten, Moran, Galun, Atzmon, Ronen, and Lipman}]{yariv2020multiview}
Yariv, L.; Kasten, Y.; Moran, D.; Galun, M.; Atzmon, M.; Ronen, B.; and Lipman, Y. 2020.
\newblock Multiview Neural Surface Reconstruction by Disentangling Geometry and Appearance.
\newblock \emph{Advances in Neural Information Processing Systems}, 33.

\bibitem[{Yu et~al.(2022)Yu, Peng, Niemeyer, Sattler, and Geiger}]{Yu2022MonoSDF}
Yu, Z.; Peng, S.; Niemeyer, M.; Sattler, T.; and Geiger, A. 2022.
\newblock {MonoSDF}: Exploring Monocular Geometric Cues for Neural Implicit Surface Reconstruction.
\newblock \emph{ArXiv}, abs/2022.00665.

\bibitem[{Yu, Sattler, and Geiger(2024)}]{yu2024gaussianopacityfieldsefficient}
Yu, Z.; Sattler, T.; and Geiger, A. 2024.
\newblock Gaussian Opacity Fields: Efficient and Compact Surface Reconstruction in Unbounded Scenes.
\newblock arXiv:2404.10772.

\bibitem[{Zhang, Liu, and Han(2024)}]{zhang2024gspull}
Zhang, W.; Liu, Y.-S.; and Han, Z. 2024.
\newblock Neural Signed Distance Function Inference through Splatting 3D Gaussians Pulled on Zero-Level Set.
\newblock In \emph{NeurIPS}.

\bibitem[{Zhang et~al.(2024)Zhang, Shi, Liu, and Han}]{zhang2024learning}
Zhang, W.; Shi, K.; Liu, Y.-S.; and Han, Z. 2024.
\newblock Learning Unsigned Distance Functions from Multi-view Images with Volume Rendering Priors.
\newblock In \emph{European Conference on Computer Vision}.

\bibitem[{Zhang et~al.(2023)Zhang, Tosi, Mattoccia, and Poggi}]{zhang2023goslam}
Zhang, Y.; Tosi, F.; Mattoccia, S.; and Poggi, M. 2023.
\newblock GO-SLAM: Global Optimization for Consistent 3D Instant Reconstruction.
\newblock In \emph{IEEE/CVF International Conference on Computer Vision}.

\bibitem[{Zhou et~al.(2023)Zhou, Ma, Li, Liu, and Han}]{zhou2023levelset}
Zhou, J.; Ma, B.; Li, S.; Liu, Y.-S.; and Han, Z. 2023.
\newblock Learning a More Continuous Zero Level Set in Unsigned Distance Fields through Level Set Projection.
\newblock In \emph{ICCV}.

\bibitem[{Zhou et~al.(2024)Zhou, Zhang, Ma, Shi, Liu, and Han}]{udiff}
Zhou, J.; Zhang, W.; Ma, B.; Shi, K.; Liu, Y.-S.; and Han, Z. 2024.
\newblock UDiFF: Generating Conditional Unsigned Distance Fields with Optimal Wavelet Diffusion.
\newblock In \emph{CVPR}.

\bibitem[{Zhou et~al.(2017)Zhou, Brown, Snavely, and Lowe}]{dblp:conf/cvpr/ZhouBSL17}
Zhou, T.; Brown, M.; Snavely, N.; and Lowe, D.~G. 2017.
\newblock Unsupervised Learning of Depth and Ego-Motion from Video.
\newblock In \emph{CVPR}, 6612--6619.

\bibitem[{Zhu et~al.(2023)Zhu, Peng, Larsson, Cui, Oswald, Geiger, and Pollefeys}]{nicerslam}
Zhu, Z.; Peng, S.; Larsson, V.; Cui, Z.; Oswald, M.~R.; Geiger, A.; and Pollefeys, M. 2023.
\newblock {NICER-SLAM:} Neural Implicit Scene Encoding for {RGB} {SLAM}.
\newblock \emph{CoRR}, abs/2302.03594.

\bibitem[{Zhu et~al.(2022)Zhu, Peng, Larsson, Xu, Bao, Cui, Oswald, and Pollefeys}]{Zhu2021NICESLAM}
Zhu, Z.; Peng, S.; Larsson, V.; Xu, W.; Bao, H.; Cui, Z.; Oswald, M.~R.; and Pollefeys, M. 2022.
\newblock NICE-SLAM: Neural Implicit Scalable Encoding for SLAM.
\newblock In \emph{{IEEE} Conference on Computer Vision and Pattern Recognition}.

\end{thebibliography}
}

\clearpage
\newpage

\end{document}